\documentclass[fleqn,10pt]{wlscirep}
\usepackage[utf8]{inputenc}
\usepackage[T1]{fontenc}
\usepackage{array}
\usepackage{multirow}

\title{Safe and Transparent Robots for \\Human-in-the-Loop Meat Processing}

\author[1,*]{Sagar Parekh}
\author[1,*]{Casey Grothoff}
\author[2]{Ryan Wright}
\author[2]{Robin White}
\author[1,+]{Dylan P. Losey}
\affil[1]{Dept. of Mechanical Engineering, Virginia Tech, Blacksburg, VA, USA.}
\affil[2]{Dept. of Animal and Poultry Science, Virginia Tech, Blacksburg, VA, USA.}
\affil[*]{Both authors contributed equally}
\affil[+]{Corresponding author can be contacted at: Dylan Losey, Goodwin Hall, 635 Prices Fork Road, Blacksburg, VA 24061, USA. Email: \texttt{losey@vt.edu}}

\begin{abstract}

Labor shortages have severely affected the meat processing sector.
Automated technology has the potential to support the meat industry, assist workers, and enhance job quality.
However, existing automation in meat processing is highly specialized, inflexible, and cost intensive.
Instead of forcing manufacturers to buy a separate device for each step of the process, our objective is to develop general-purpose robotic systems that work alongside humans to perform multiple meat processing tasks.
Through a recently conducted survey of industry experts, we identified two main challenges associated with integrating these collaborative robots alongside human workers.
First, there must be measures to ensure the safety of human coworkers; second, the coworkers need to understand what the robot is doing.
This paper addresses both challenges by introducing a safety and transparency framework for general-purpose meat processing robots.
For \textit{safety}, we implement a hand-detection system that continuously monitors nearby humans.
This system can halt the robot in situations where the human comes into close proximity of the operating robot.
We also develop an instrumented knife equipped with a force sensor that can differentiate contact between objects such as meat, bone, or fixtures.
For \textit{transparency}, we introduce a method that detects the robot's uncertainty about its performance and uses an LED interface to communicate that uncertainty to the human. 
Additionally, we design a graphical interface that displays the robot's plans and allows the human to provide feedback on the planned cut.
Overall, our framework can ensure safe operation while keeping human workers in-the-loop about the robot's actions.
To demonstrate the capability of our system, we conduct a user study with robotics and meat processing experts, and compare our framework with industry-standard mechanisms.
Subjective results suggest that experts prefer our safety and transparency framework over existing alternatives.
Further, a pre- and post-demonstration survey indicates that --- after interacting with our  framework --- experts are less hesitant to work alongside general-purpose robot arms in meat processing facilities.

\p{Keywords} Robotics, Meat Processing, Human-Robot Interaction, Industrial Automation 

\end{abstract}

\usepackage[numbers]{natbib}

\newcommand{\subsubsubsection}[1]{\paragraph{#1}\mbox{}\\}
\newcommand{\p}[1]{\smallskip \noindent \textbf{{#1}.}}
\newcommand{\eq}[1]{Equation~(\ref{eq:#1})}
\newcommand{\fig}[1]{Figure~\ref{fig:#1}}

\begin{document}

\flushbottom
\maketitle
%
%
\thispagestyle{empty}

\section{Introduction}

Global meat production in $2024$ reached $379$ million tonnes according to the Food and Agriculture Organization (FAO) of the United Nations \citep{fao}. This figure comprises $150$ million tonnes of poultry, $78.7$ million tonnes of bovine meat, and $125.1$ million tonnes of pig meat, with poultry and bovine meat production experiencing year-on-year growth of $2.6 \%$ and $2.8 \%$, respectively. The growth in the overall meat production has seen a steady increase in the past years --- $1.4 \%$ in $2024$, $1.5 \%$ in $2023$, and $1.6 \%$ in $2022$. The continued increase in global meat production underscores the rising demand for meat products worldwide. According to FAO projections, global meat consumption is expected to grow by $14 \%$ by $2030$, driven by population growth, urbanization, and rising incomes, particularly in developing regions. However, this growing demand poses significant challenges for the meat processing industry, which is struggling with persistent labor shortages. This problem has been exacerbated by the recent COVID-19 pandemic \citep{taylor2020,bir2021impact}. Attracting and retaining skilled workers has proven difficult, due in part to the negative perception of meat processing jobs, which are often viewed as physically demanding and low-satisfaction careers \citep{victor2016slaughtering,gaston2012meatpacking}, and associated with elevated risks of injury and post-traumatic stress disorders \citep{pastrana2023slaughterhouse,slade2023psychological}.

To meet increasing demand while addressing labor challenges, there is a pressing need to develop advanced, scalable, and sustainable manufacturing systems for meat processing. Automation has emerged as a potential solution for supporting more streamlined and reliable supply chains, ensuring consistent and rigorous animal welfare standards, and maintaining reliability in meat product availability, quality,
and price \citep{aly2023robotics,madsen2006automation}. However, automation has largely been limited to large-scale operations that can afford the high costs associated with highly specialized machinery \citep{wakholi2021economic,kim2021economic,hinrichsen2010manufacturing}. For automation to be viable industry-wide, new manufacturing technologies must be both flexible and scalable, enabling small and medium-scale processors to adopt them effectively. In this context, collaborative robots (cobots), i.e., robots designed to work alongside humans, have gained attention as a promising next step \citep{kakade2023applications,keshvarparast2024collaborative,parekh2023learning,puttero2025collaborative}. Cobots not only enhance efficiency and reduce production costs \citep{romanov2022towards,sahan2023role}, but also alleviate the physical and psychological burden on human workers by taking over dangerous, repetitive, and low-satisfaction tasks \citep{fournier2024human,bouillet2025effects,su2024exploring}. But the major advantage of cobots lies in their versatility --- a single cobot can perform multiple meat processing operations as well as assist human workers in their tasks. Due to this capability cobots can be a cost-effective alternative to specialized automation for small and medium scale industries.

\begin{figure}
    \centering
    \includegraphics[width=0.75\linewidth]{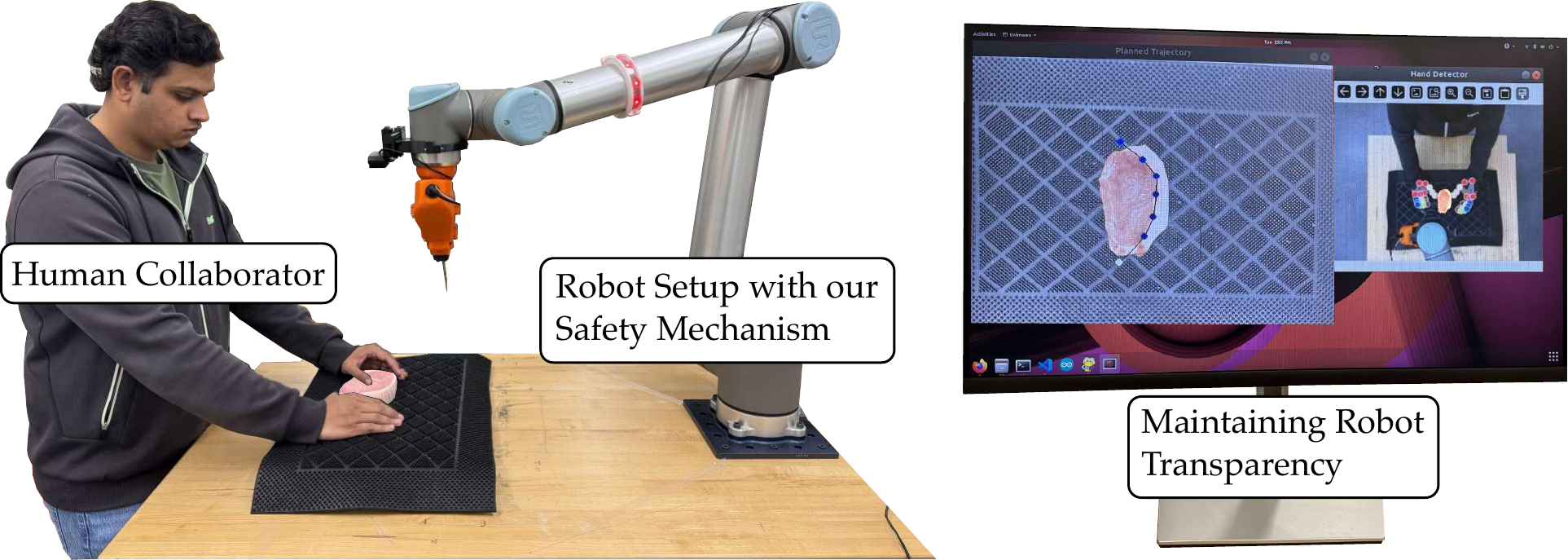}
    \caption{Our framework for ensuring safety, maintaining transparency, and integrating human feedback into collaborative robotic systems. The robot works in close proximity to human workers, typically working on processing the same meat product. We propose an instrumented knife and a human hand monitoring system for anticipating in real-time the chances of unwanted interaction between the robot and the human or the environment. We also leverage an LED interface for communication anomalies to the human coworkers. Further, we develop a graphical interface that displays the robot's planning process while simultaneously allowing the human to modify the robot's planned cut.}
    \label{fig:front}
\end{figure}

In our previous work \citep{wright2024safely} we demonstrated the feasibility of deploying a collaborative robot for meat processing tasks, showcasing its ability to adapt to varied meat types and cuts. The robot consistently produced products within industry allowable margins of error. Furthermore, a survey of meat processing professionals in the United States revealed that participants found the robot-generated cuts to be satisfactory. Notably, experts expressed a preference for collaborative automation, where a human remains part of the planning process, over fully automated operation. 
While these results showcase the engineering feasibility of collaborative robots for meat processing application, there are still challenges pertaining to integrating such a technology on the factory floor.
First and foremost, the challenge with adopting cobots into industry lies in the safety of human workers.
Since cobots share a workspace with humans, often interacting with each other to complete tasks, it is critical to ensure safety during the operation.
Secondly, the success of collaborative robots is contingent on the human coworker's trust in the robot's capabilities \citep{xing2018smart,mutlu2016cognitive} --- a black-box robot that does not share its plans with the human is deemed less trustworthy by humans.

In this paper we advance our long-term goal of developing versatile cobots for meat processing applications.
We recently surveyed industry experts to learn their perceptions of automation and identify targets for future work \citep{wright2025survey}.
Based on the survey responses, we here identify and address three specific considerations for deploying collaborative robots in meat processing facilities: \textit{safety}, \textit{transparency}, and \textit{feedback integration}.
Intuitively, the former encompasses the safety issue associated with cobots that must operate in close proximity to humans \citep{haddadin2016physical,lasota2017survey}, and the latter two help improve the human's trust in the robotic system by introducing a two-way communication channel between the human and the robot \citep{habibian2025survey,yang2017evaluating,yao2025robot} (see \fig{front}).
For safety, we propose real-time sensing and fail-safe mechanisms that prevent any collision between the robot and human or environment.
To this end, we leverage a vision-based human hand monitoring system and an instrumented knife.
Additionally, we realize that the people working alongside the cobot may not necessarily be experts in robotics.
Consequently, it is important that the robot maintains a certain level of transparency by clearly conveying its goals, intentions, and uncertainties to the human worker.
We design two interfaces: a graphical interface to display the robot's intended cutting plan in a way that is interpretable by human coworkers, and an LED interface to promptly notify human supervisors in case of failures.
The graphical interface also integrates the human into the robot's planning process by enabling them to provide feedback on the robot's planned cuts.
Specifically, if the proposed cutting plan of the robot is unsatisfactory to the worker, they can easily make corrections to the cutting plan using an intuitive click-and-drag interface.
Overall, our framework constitutes components that satisfy the three aforementioned considerations for effectively adopting cobots into the workplace. To evaluate the usefulness of our proposed framework we conduct a proof-of-concept demonstration involving $20$ experts recruited from both the meat processing and robotics communities.

\section{Methods}\label{sec:method}

Our long-term goal is to develop multi-purpose collaborative robots that can be integrated into meat processing industries.
As a step towards this goal, we introduce a framework designed to address the key challenges commonly encountered with collaborative robots.
Informed by preliminary surveys with meat industry experts, the framework focuses on three main aspects: safety, transparency, and feedback integration.
Here we discuss each component of the framework and describe the experimental methodology used for its validation.
In Section \ref{M1} we outline our multi-sensor setup aimed at ensuring safe execution of tasks.
In Section \ref{M2} we describe our approach for enhancing robot transparency; this framework enables robots to effectively communicate their objectives and plans to human collaborators.
Finally, in Section \ref{M3} we introduce our mechanism for getting feedback from humans and seamlessly integrating that feedback into the robot's plan.

\subsection{Safety: Monitoring Robot Interaction}\label{M1}
Collaborative robots are designed to operate in close physical proximity to humans, and in order for human coworkers to confidently collaborate with such robots it is important to ensure human safety. 
In typical meat processing tasks, human workers would share the same workspace as the robot.
Humans may place meat in or remove it from the robot's workspace, complete simultaneous tasks on the same carcass, or operate in sequence alongside the robot.
Such scenarios introduce the risk of unintended physical contact between the robot and the human.
Beyond the safety of human workers, operational safety also requires preventing the robot from interacting with its environment in undesirable ways. 
For example, the cutting tool used by the robot must avoid contact with objects like the fixtures used to hold the meat. 
Such unintended contact could damage the equipment or introduce additional hazards due to breakage.
In physical human-robot collaboration, safety is often conceptualized in terms of collision avoidance between humans and robots \citep{de2008atlas}. 
A straightforward and intuitive approach to minimizing this risk is to stop the robot’s motion immediately when a human enters its workspace or when an unwanted interaction is detected. 
This necessitates accurately detecting humans as well as identifying unsafe interactions. 
To address this, we introduce a safety mechanism that leverages a vision-based hand detection system to continuously track human movement relative to the robot and an instrumented knife capable of distinguishing between contacts with different objects.
The former can identify when a human enters the robot’s workspace and the latter can identify if the knife is in contact with meat, as intended, or with unintended objects such as the table or other hard surfaces. 
The technical details of our vision system and the instrumented knife design are discussed in the following sections.

\subsubsection{Vision-Based Hand Monitoring System}\label{M11}
At the start of the collaborative task the human coworker places the meat at a designated location in front of the robot. 
We define this fixed location for meat placement and specify a bounded region $W_{op}$ around its location as the robot's operational workspace.
To ensure safety, we start by kinematically constraining the robot to strictly operate within $W_{op}$ \citep{wright2024safely}.
This constraint reduces the risk of unintended physical contact between humans and the robot.
However, there are necessary instances when the human must enter $W_{op}$, such as during the placement of meat, or when the human is helping to process that same meat. 
These instances introduce a potential for contact or collision, and simply using a bounding box will not be sufficient.
To mitigate this risk, we propose a vision-based safety mechanism.
Specifically, an RGB camera is mounted above the robot to provide a top-down view of $W_{op}$ as well as the surrounding region, as shown in \fig{safety} (Left).
In the image frame of the camera, we define two safety zones: first, we define a safe zone which is located farther away from the robot, and which the robot cannot reach on account of its kinematic constraints; second, we define a warning zone which includes the operational workspace $W_{op}$ of the robot and its immediate surrounding.
These zones are illustrated in \fig{safety} (Middle-left).
The vision system continuously monitors for humans placing their hands within these zones.
To detect human presence, we use Google MediaPipe’s hand landmark detection model \citep{mediapipe}, which identifies a set of $21$ hand landmarks $\{p_i\}, \, i = 1, 2, \cdots, 21$ in the $2$D image space.
The detection model operates on each image frame in the camera stream, enabling real-time monitoring of human hand positions.
When a detected landmark lies within the warning zone in the image frame, a safety trigger immediately stops the robot's movement.
This stopping mechanism greatly reduces the risk of any potential contact between the robot (or the knife it holds) and the human.

\subsubsection{Instrumented Knife}\label{M12}
\begin{figure}
    \centering
    \includegraphics[width=1\linewidth]{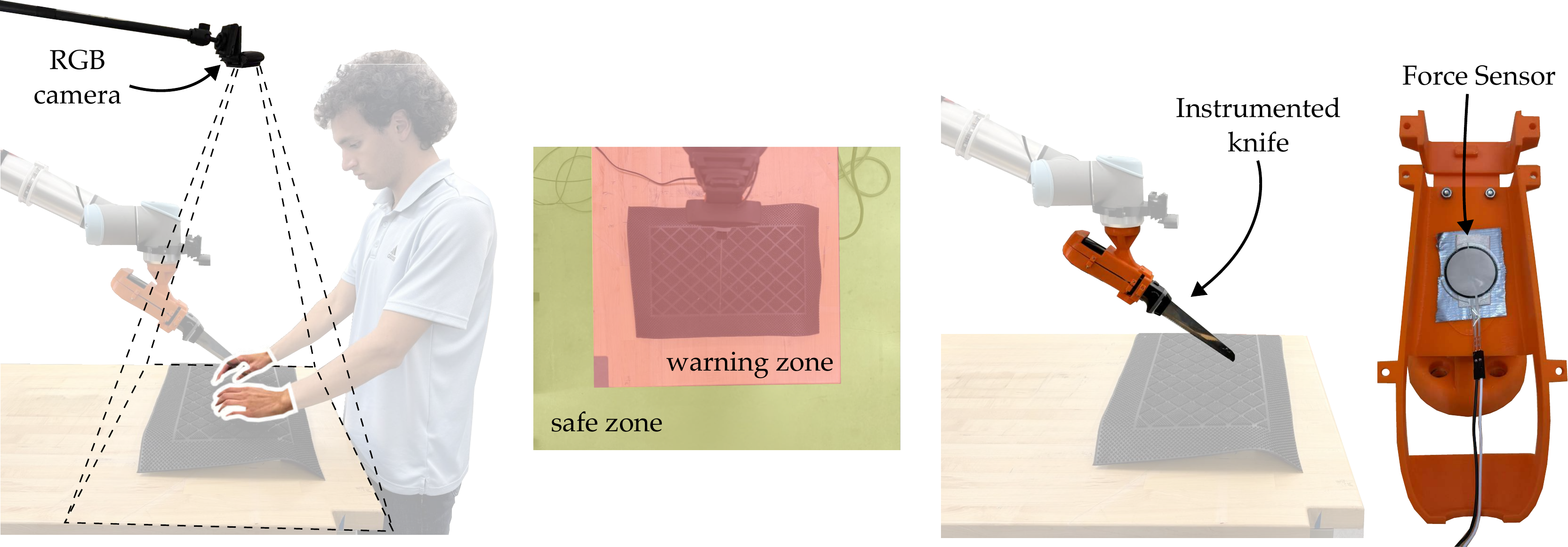}
    \caption{Safety mechanism of our robot framework. (Left) The vision based hand monitoring system. A camera is mounted above the robot arm to obtain a top-down view of the robot's workspace $W_{op}$ and the surrounding environment. The camera constantly monitors the region to detect human hand entering the workspace. (Middle-left) We divide the region into two zones. A safe zone where the robot cannot reach on account of the kinematic constraints imposed on its movement and a warning zone where the robot operates. We monitor is the human hand is in these regions to ensure that the robot can proactively pause its movement to reduce the possibility of making contact with the human. (Middle-right) The instrumented knife that is attached to the robot using a custom-designed mount equipped with a force sensor to constantly monitor the resistance encountered by the knife. Intuitively, if the knife makes contact with any object other than meat, which is likely to be harder, it will experience a larger resistance which can be used as a distinguishing factor for the contact. (Right) The cross-section of the custom mount which encases a force sensor.}
    \label{fig:safety}
\end{figure}

In addition to avoiding contact between human and robot, it is important that the robot’s end-effector --- specifically the cutting tool --- interacts only with the intended target, i.e., the meat. 
Any contact with undesirable objects such as the fixtures holding the meat or bones must be avoided.
Such undesirable contact can lead to operational hazards as the equipment can get damaged or the knife blade can break.
To achieve this, we introduce an instrumented knife system capable of detecting and differentiating contacts with varied objects.
We acknowledge that we developed an instrumented knife equipped with proximity sensors in \citep{wright2024safely} for detecting contact, however this knife was not capable of discriminating between contacts with different objects.
While that knife was a step towards avoiding contact with humans, the instrumented knife we present in this paper is designed to ensure operational safety by avoiding contact with objects other than meat.
We employ a KitchenAid electric carving knife mounted on a robot arm using a custom-designed 3D-printed mount, illustrated in \fig{safety} (Middle-right).
The mount holds the knife at a fixed inclination angle, optimized to ensure effective cutting and consistent contact with a force-detection module.
To monitor contact forces during operation, we integrate a FlexiForce sensor with a force range of $0$-$25$lbs \citep{flexiforce}, strategically placed so that the knife exerts pressure on it when in use, as shown in \fig{safety} (Right). 
The sensor measures the normal force exerted by the knife onto the sensor, which correlates with the resistance encountered by the knife during cutting.
To differentiate between normal cutting interactions and undesired contacts, we experimentally determine the maximum force encountered during normal operation.
Based on this maximum force magnitude we set a force threshold with a safety margin to distinguish contacts with hard objects (e.g., table, bone) and with soft objects (like meat).
During operation, the instrumented knife continuously monitors the normal force applied by the knife, and if at any point the normal force exceeds the pre-determined threshold, an emergency stop is immediately triggered.
This stop signal is sent to the robot's low-level controller to stop its movement, avoiding any undesirable damage between the equipment and the environment.
By incorporating this force-based feedback loop into the overall safety protocol, we ensure that the robot is not only aware of the human coworker’s location but is also capable of detecting and responding to unwanted physical contact in its workspace. 

\subsection{Transparency: Communicating Robot Plans to Humans}\label{M2}
So far, we have introduced a dual-layer safety mechanism designed to ensure safer human-robot interaction and reduce the risk of equipment damage.
This mechanism includes kinematic constraints and real-time monitoring during robot operation.
The next critical aspect in ensuring effective human-robot collaboration is robot transparency.
Transparency refers to the robot's ability to communicate its intentions, uncertainties, and unexpected events to the human partner \citep{winfield2021ieee}.
In our context, this includes $1$) calculating and communicating the uncertainty in execution, $2$) alerting the human in case of anomalies, and $3$) conveying the planned cutting trajectory.
To support such transparency, we propose a communication framework that is composed of a mechanism for estimating the uncertainty about previously executed cuts, an LED-based signaling interface for real-time feedback during task execution, and an interactive graphical interface for displaying the robot's future cuts.

\subsubsection{Detecting Uncertainty of Cuts}\label{M21}
We use the planning procedure developed in our previous work \citep{wright2024safely}.
While this enables the robot to autonomously plan and precisely execute the cuts, a vision-based system lacks the ability to identify bones that might be within the meat.
This can lead to potential errors as the knife may unexpectedly come into contact with the bone, causing the cutting trajectory to deviate from the originally planned path.
To handle such instances, it is essential for the robot to assess whether it actually completed the desired cut.
A direct approach would involve visually comparing the state of the meat before and after the cut to determine the cut’s accuracy. 
However, this form of visual evaluation is highly challenging due to the deformable nature of meat, lighting conditions, and other confounding visual factors.
Instead, we propose an indirect --- yet effective --- proxy measure for evaluating the robot’s confidence in its cutting performance. 
Specifically, we hypothesize that the displacement of the meat on the cutting surface can serve as an indicator of cutting success. 
The underlying intuition is that a successful cut, free from any interference like contact with bone, would result in minimal or no movement of the meat. 
In contrast, if the knife encounters resistance (e.g., hitting a bone), it may fail to cleanly complete the cut and instead drag the meat along, causing noticeable displacement.
To quantify this displacement, we first use the image of the meat captured during the planning phase. 
From this image, we segment the meat region and fit the smallest possible bounding box that fully encloses the segmented area. 
The four corners of this bounding box, expressed in pixel coordinates, are recorded as the initial location of the meat, denoted by $p_{\text{initial}} = \{(x_{1} y_{1}), (x_2, y_2), (x_3, y_3), (x_4, y_4)\}$.
Following human approval, the robot executes the planned cutting trajectory.
After completing the cut, the robot returns to its initial observation position and captures another image of the meat.
Using the same segmentation and bounding box fitting procedure, we calculate the post-cut location $p_{\text{post}} = \{(x_{1}, y_{1}), (x_2, y_2), (x_3, y_3), (x_4, y_4)\}$.
We then estimate the displacement as the mean Euclidean distance between the bounding boxes pre- and post-cut:
\begin{equation*}
    d = \frac{1}{4} \sum_{i=1}^{4} \left| p_{\text{post}}^i - p_{\text{initial}}^i \right|
\end{equation*}
where $p^i = (x_i, y_i)$ from the initial and post-cut locations.
Since we compare the four corners of the bounding box, the displacement also accounts for any rotation of the meat.
To convert this displacement into the robot's uncertainty, we define an uncertainty function $\Psi(d)$ that maps the displacement to a normalized confidence score.
This function outputs a value between $0$ and $1$ which can be directly interpreted as the robot's uncertainty about the cut it made.
Further, this function allows easily changing the sensitivity to the displacement.
$\Psi(d)$ is defined as
\begin{equation}
    \Psi(d) = \frac{e^{\beta d} - e^{-\beta d}}{e^{\beta d} + e^{-\beta d}}
\end{equation}\label{eq:uncertainty}
Here $\beta > 0$ is a tunable parameter that controls how sensitive the detection system is to various displacements. 
A larger value of $\beta$ implies that even small displacements will yield higher uncertainty values, reflecting stricter tolerance for error.
Conversely, a smaller value of $\beta$ allows greater tolerance to the displacement.
This displacement-based approach enables the robot to reason about the success of its cutting action without requiring high-fidelity visual inspection. 

\subsubsection{LED Interface}\label{M22}
Until now, we have outlined various mechanisms by which the robot can detect anomalies in its operation.
In Section \ref{M11}, we introduced a vision-based system for monitoring the human's presence in the robot's workspace.
Section \ref{M12} described a method for detecting unintended interactions between the knife and other objects, and in the preceding section, we presented an approach for quantifying the robot’s performance uncertainty during task execution.
The first two mechanisms are designed to enable the robot to either pause or completely stop its operation when a safety-compromising situation is detected --- such as the presence of a human in the workspace or undesirable interaction of the knife with the environment.
The third mechanism allows the robot to assess whether the quality of its cut meets the expected standard.
However, the efficacy of our framework relies not only on these mechanisms' abilities to detect these anomalies but also on the robot's capability to communicate clearly and immediately to the human collaborator.
This requirement is particularly important in realistic industrial settings, where it is not feasible to expect a human operator to maintain constant supervision over the robot.
Instead, we assume a collaborative environment where the robot performs its tasks semi-autonomously, while the human worker simultaneously focuses on a separate task.
For instance, there could be multiple robots operating in the processing plant and a human partner may have to transport and orient the meat at each work station.
It is not feasible to assume that a human coworker is going to careful monitor the robot's cutting trajectory every time.
Consequently, it becomes critical for the robot to actively gain the human’s attention whenever an intervention is needed.

\begin{figure}
    \centering
    \includegraphics[width=1\linewidth]{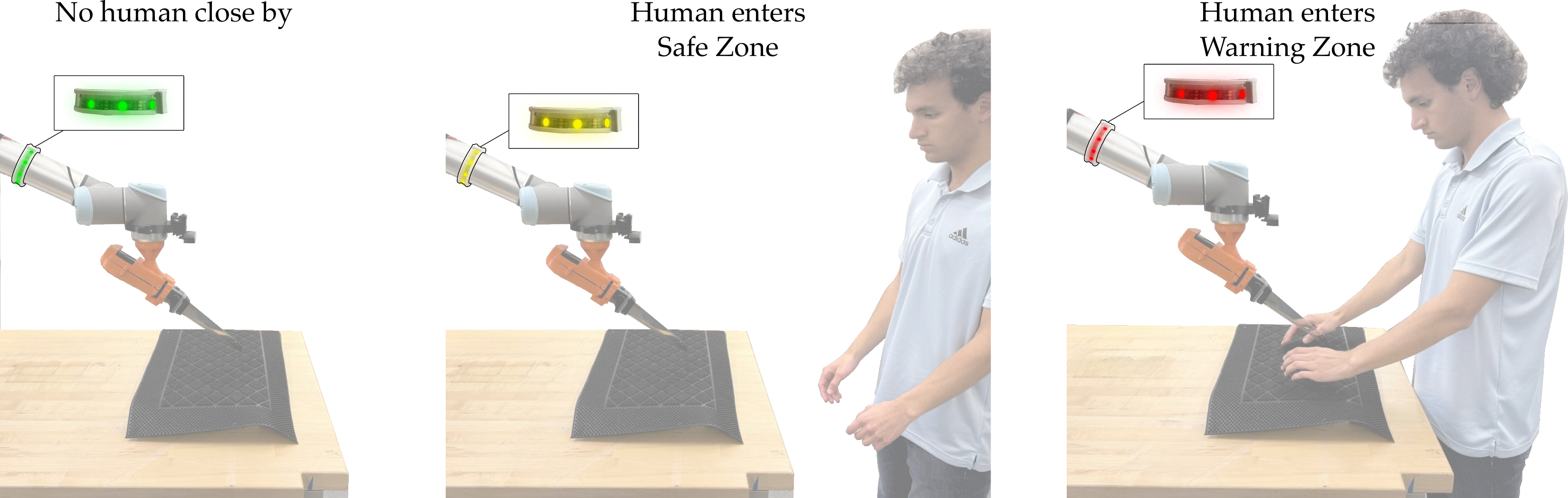}
    \caption{The LED interface used as a communication channel between robot and human. Here, we demonstrate the LED interface communicating the different situations which might be encountered during the robot's execution of the cuts. (Left) When the human is farther away from the robot, it is in normal operation mode and the LED lights are green. (Middle) When the robot approaches the robot, entering the safe zone, the light turn yellow. This helps communicate to the human that they are approaching the robot's workspace where they can accidentally make contact with the cutting tool. (Right) The LED interface turns red when the human is inside the robot's workspace or the warning zone.}
    \label{fig:led}
\end{figure}

To this end, we designed and implemented a light-based signaling interface using LED indicators, which serves as a simple yet effective communication channel between the robot and the human.
Lights are also well-suited for noisy factory settings where other modalities (like sound) may be impractical.
The LED interface uses RGB lights which can change color to indicate different situations.
Specifically, the LED turns green when the robot is in normal operation mode, i.e., the operation is safe and the robot does not detect any anomalies.
The LED turns red to convey failure cases such as when the robot stops due to the detection of a human entering the workspace or because the knife has made contact with an unintended object, see \fig{led}.
For instance, in the case of humans entering the robot's workspace $W_{op}$ the lights turn yellow when the human hand is detected to be in the safe zone and red when the human hand is detected to be in the warning zone.
Finally, if the uncertainty function at the end of the cutting task exceeds a predefined threshold --- indicating that the robot lacks confidence in the quality of the cut --- the LED also turns red, alerting the human to inspect the result and provide assistance if needed.

\subsubsection{Communicating Robot Plans}\label{M23}
In our previous work \citep{wright2024safely} we developed a vision-based planning system that autonomously generates cutting trajectories for various meat-processing tasks. 
The planning system uses an RGB camera mounted on the robot’s end-effector to capture an image of the meat placed by the human coworker in the workspace.
The system processes the image to perform segmentation, identifying each pixel as belonging to either meat or fat using predefined RGB color thresholds.
The boundaries of the segmented meat and fat form the basis for the planned cutting trajectory.
The planned trajectory in the image frame is a sequence of $(x, y)$ pixel coordinates that marks the cut the knife should make on the meat.
This $2$D path is then transformed to a $3$D trajectory using our calibration procedure which serves as the commanded path for the robot knife edge during execution.

\begin{figure}
    \centering
    \includegraphics[width=1\linewidth]{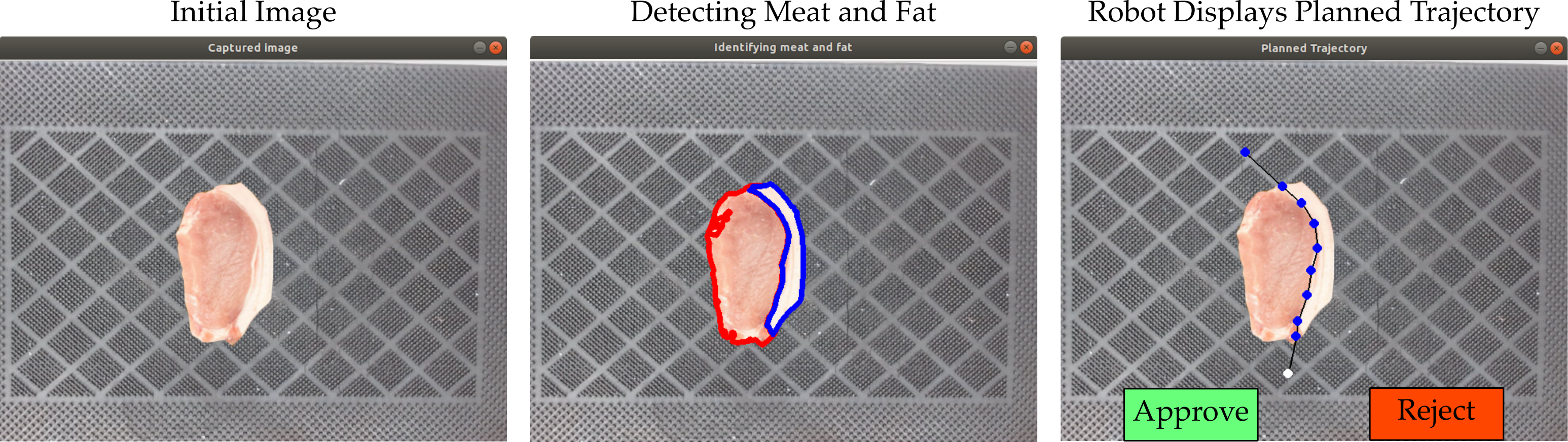}
    \caption{The graphical interface for maintaining transparency about the robot's planning procedure. The robot displays the sequence of steps for how it will cut this piece of meat. (Left) A camera is mounted on the robot arm which captures the image of the meat placed on the cutting surface. (Middle) For an example task where the robot must trim excess fat off the meat, the system uses thresholds on the RGB values for meat and fat to identify and segment these parts in the image. Once the meat and fat are segmented, a $2$D trajectory, which is a sequence of points in the pixel coordinates, is planned by the robot. (Right) The planned trajectory is displayed to the human coworker by overlaying it on the meat image. This interface includes buttons which allow the human to either accept or reject the cutting plan. If needed, the human can drag any point in the trajectory to make adjustments to the cut. Additionally, they can also add or remove points in the trajectory. The updated cut is then executed instead of the original robot plan.}
    \label{fig:communication}
\end{figure}

While previously this automated planning process was a black-box (and the human would only see the trajectory during execution), we now add transparency to the planning procedure to ensure that the human partner remains informed and stays in control.
Specifically, we design a graphical interface that displays the robot’s intended plan before execution. 
The procedure starts with the human partner placing the meat in the robot's calibrated workspace.
The robot captures an RGB image and computes the $2$D trajectory in pixel coordinates.
This $2$D trajectory is overlayed onto the captured image of the meat, which is then displayed on the graphical interface.
The robot waits for the human to provide explicit approval via the UI button before execution.
Upon approval the robot transforms the trajectory from pixel coordinates to the robot coordinates and performs the cut.
We hypothesize that this visual confirmation step will increase the human’s trust and understanding of the robot’s behavior.
An illustration of the interface is shown in \fig{communication}, where the graphical interface displays the entire planning process including  detected contours for fat and meat (in the middle), and the path the robot plans for cutting (on the right).

\subsection{Feedback: Integrating Human Inputs into the Robot's Plan}\label{M3}
In the previous section we introduced the key components of our framework that enable the robot to maintain transparency and communicate effectively with its human partner, particularly in scenarios where task execution may result in suboptimal or unfavorable outcomes. 
While such transparency is critical for collaboration, effective communication alone is insufficient to ensure robust and adaptive human-robot interaction.
The robot's planning process is performed autonomously using digital image processing techniques applied to an image of the meat.
However, this process is inherently susceptible to a wide range of variability both in the input and the environment.
Factors such as meat size, shape, and color along with environmental artifacts such as lighting conditions, shadows, and reflections introduce uncertainty in robot perception, which can ultimately results in suboptimal planned cuts.
Although the graphical interface described previously enables the human partner to review and reject an unsatisfactory trajectory, it still relies on the same autonomous planner to regenerate its new trajectory.

Given that humans possess superior high-level reasoning capabilities and contextual understanding, they are often able to visually assess the meat and mentally plan a more suitable cut with minimal effort.
Thus, integrating human inputs into the planning loop is not only desirable but also essential for achieving efficient and task-aligned execution.
The human coworker must be able to not just approve or reject the robot-generated plans, but to directly revise them to better reflect their preferences and domain expertise.
To this end, we modify the existing graphical interface so that the human can modify and manually specify the planned cuts.
If the human thinks the trajectory is unsatisfactory, they are provided with a tool to intuitively revise it.
Specifically, the robot's cutting trajectory is represented as a sequence of waypoints --- discrete spatial robot coordinates that define the path of the robot's end-effector and mounted knife. 
The robot follows a linear path between two consecutive waypoints, and the sequence is optimized to minimize the number of waypoints while still accurately approximating the curvature of the meat.
The interface allows the human partner to interact with the planned trajectory through a simple click-and-drag operation: they can add a new waypoint in the sequence by clicking on the image, they can remove a waypoint by clicking on it, and they can modify a waypoint's position by dragging it to new locations.
Once the human has completed their modifications and is satisfied with the revised plan, they can approve the trajectory. 
The robot then replaces its original trajectory with the updated one and proceeds with execution accordingly.
This human-in-the-loop planning mechanism enables the robot to incorporate human feedback efficiently and intuitively.

\subsection{Experimental Protocol}
We have discussed the individual components that constitute our proposed safety and transparency framework for meat processing application.
In this section we discuss our overall robotic setup, outlining how each component is incorporated in our pipeline.
Furthermore --- to test if our framework can reliably ensure safety and transparency during operation --- we conduct a series of experiments as well as demonstrations.
We also detail the procedure we follow for validating our framework.

\subsubsection{Robot Arm Setup}

\begin{figure}
    \centering
    \includegraphics[width=0.8\linewidth]{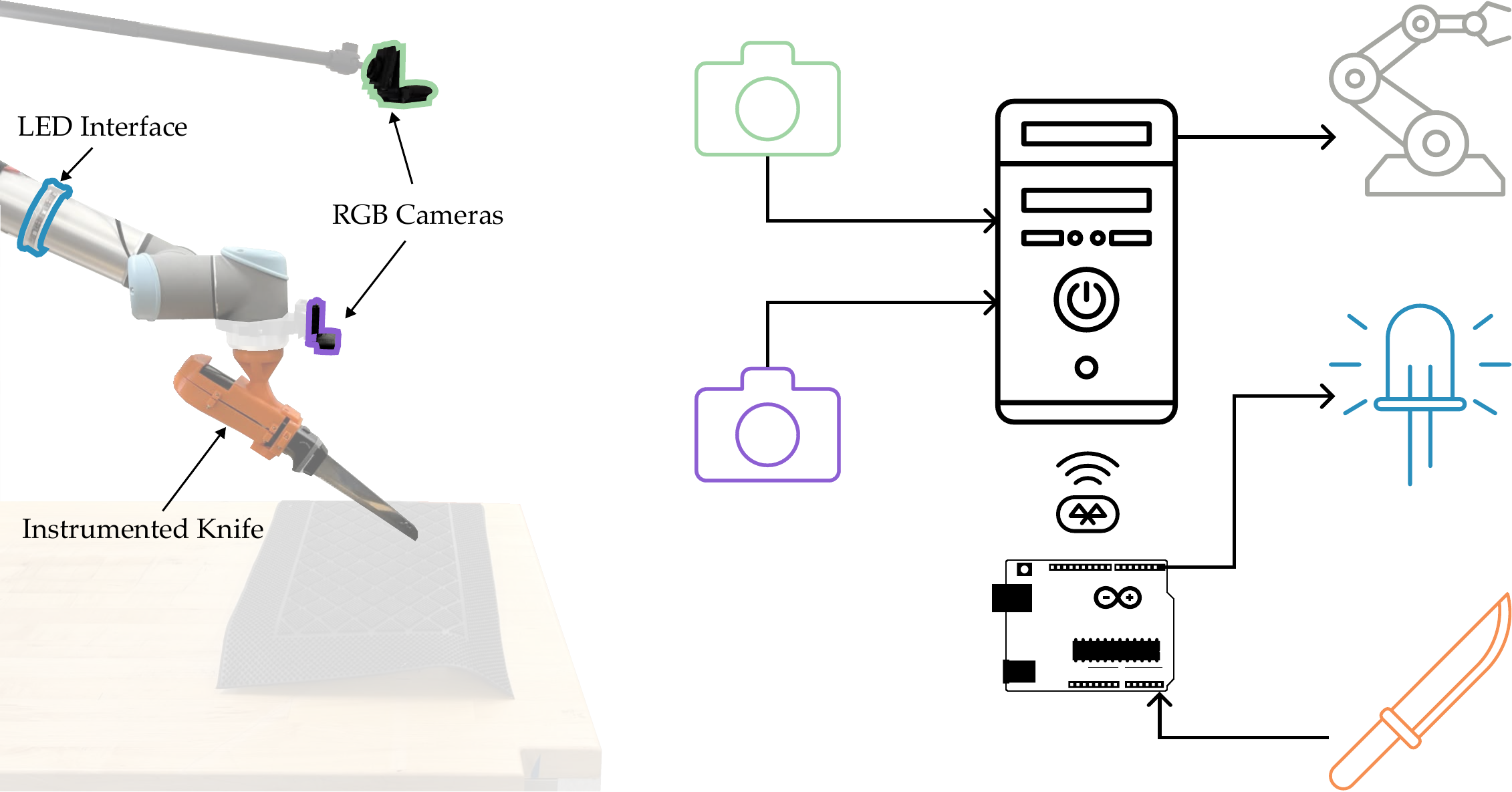}
    \caption{Robot arm setup used in our framework. (Left) The setup includes two RGB cameras, one mounted on the robot for planning cutting trajectories and another mounted overhead for hand monitoring, our instrumented knife, an LED interface controlled by a micro-controller, and a UR10 robot. (Right) The schematic  diagram shows the communication flow of our framework. The two RGB cameras are controlled by the intel NUC which also controls the robot. The image from the robot-mounted camera is used for planning the robot cutting trajectory and is executed by the robot. The micro-controller monitors the force reading from the instrumented knife and controls the LED interface. This micro-controller communicates with the NUC over bluetooth and whenever an undesirable contact is detected by the knife it sends a stopping signal to the NUC which immediately stops the robot's movement.}
    \label{fig:overall_setup}
\end{figure}

We perform our experiments with a $6$ DoF UR$10$ robot arm developed by Universal Robots.
This robot is commercially used in industrial applications for various tasks including assembly, painting, and welding \citep{ur10}.
The robot is compliant with the ISO $10218-1$ regulations for collaborative robots as mentioned on the manufacturer website \citep{ur10_2}.
We used two RGB webcams (Logitech C290) in our framework; one webcam is mounted on the robot arm using a custom $3$D-printed mount and is used for localizing the meat in the robot's trajectory planning process, and the second webcam is mounted above the robot using a store-bought mount and is used for safety monitoring.
We have a computer, a NUC with $8^{th}$ gen Intel Core i$7$-$8559$U processor and Intel Iris Plus Graphics 655, connected to the robot and the two webcams for handling the processing, i.e., planning and executing the cuts as well as monitoring the human.
For cutting operations we used a KitchenAid carving knife that was mounted on the robot using our custom $3$D-printed mount as shown in \fig{safety} (Right).
The force sensor embedded inside the knife mount is operated using an Arduino Uno R$3$ micro controller which constantly reads the sensor measurements and controls the LED interface.
The overall setup is shown in \fig{overall_setup}.
We emphasize that while we utilize these specific combination of robot and devices, our comprehensive framework for safety, transparency, and feedback integration can be directly extended to other camera, robot, and knife combinations.

The camera mounted on the robot arm captures images of the meat once that meat is placed on the designated location in the robot's workspace.
These images are then sent to the connected computer for processing; here the system uses thresholds on RGB values to detect the meat and the fat. 
Depending on the task at hand, the desired trajectory for the cut is computed in pixel coordinates.
For example, if the task is to cut the meat into slices the program fits equally spaced parallel cuts to slice the meat into $n$ strips.
If the task is to trim the excess fat, the program fits a cutting trajectory along the intersection of the detected fat and meat.
Before the robot can start executing the cut, we need to convert the cutting trajectory from pixel coordinates into a path in robot coordinates that the robot can follow to make the right cut.

The computer connected to the robot runs a high-level control program which calculates the desired velocity for the robot based on the planned trajectory.
The velocity commands are sent to the low-level controller of the robot which executes the motion with precise accuracy at a frequency of $500$Hz.
Simultaneously, the computer runs a safety program at a frequency of $60$Hz that uses the overhead mounted camera to monitor the human's movement.
As soon as a human is detected entering the boundary of the robot's workspace $W_{op}$, i.e., when they enter the safety zones, the safety program communicates this information to the high-level control program.
Then, the control program turns the velocity for the robot to zero, essentially pausing the robot's movement.
The control program keeps the robot's velocity at zero until it receives a safety flag from the safety program indicating that the human is out of $W_{op}$.
At the same time, the safety program also communicates to the micro controller whether the human is in the safe zone or the warning zone.
Depending on which zone the human is in, the micro controller operates the LED interface to flash yellow light, indicating the human is safe but approaching the robot, or red light, indicating the human is too close to the robot.
Additionally, the micro controller continuously reads the force sensor measurements. 
When the force value crosses the safety threshold it does two things: communicates to the computer that the knife is interacting with an object other then the meat and flashes the LED interface red.
Upon receiving this information the control program immediately stops the robot and ends the execution.

\subsubsection{Evaluation Procedure}\label{sec:evaluation_procedure}
In this subsection, we outline the experimental procedure for validating the proposed collaborative robot framework designed for meat processing tasks.
The evaluation is divided into two main phases: $1$) individual component testing, where the different subsystems of the framework are evaluated under controlled conditions, and $2$) full system demonstration, where the integrated framework is evaluated through in-person demonstrations for experts from meat processing and robotics domains.
The goal of this two tiered evaluation methodology is to not only verify the robustness of our proposed framework but also obtain qualitative feedback on the system's usability from potential users.

\subsubsubsection{Evaluating Individual Components}
The first phase of the experimental procedure focuses on evaluating three main components of the framework, namely the human monitoring system, the instrumented knife, and the uncertainty detection module.
Each component is tested under controlled settings that represent real-life meat processing workflow.
It is important to note that we do not include experiments for the graphical interface.
The interface mainly serves two purposes: to display the planned robot cut to the human operator, and update the robot plan based on human feedback.
These functionalities (overlaying the cut on the image and tracking the mouse pointer) do not influence the robot's performance.
The only potential source of error stems from miscalibration.
However, experimental results testing our calibration have already been presented in our previous work \citep{wright2024safely}.
Therefore, we omit interface-specific experiments in this work.

\p{Human Monitoring System}
To evaluate this component, we recruit $5$ participants to interact with our robot setup as it simulates a typical meat processing task.
To ensure safety of the participants, we do not mount the knife onto the robot or cut actual meat.
The objective of this experiment is not to evaluate the cutting performance of the robot, but to test the reliability of our human monitoring system.
During our experiments, the participants repeatedly interact with the robot emulating a meat cutting scenario where they place the meat in front of the robot, followed by the robot executing a preplanned trajectory.
During operation, the participants are instructed to intermittently enter the robot's workspace to re-position the meat.
This process is repeated ten times to ensure consistency resulting in a total of fifty trials.
To evaluate performance, we measure four quantitative metrics: first, we measure the detection accuracy defined as the percentage of true positives (correctly detecting when human enters the workspace); second, we measure the precision defined as the proportion of true positives to the total predicted positives; third, we measure the false negatives (incorrectly detecting human entering the workspace); and finally, we measure the latency of the robot, i.e., how much time it takes for the robot to stop its motion.
For the latency metric, we record the precise timestamps at two critical events --- when the human is first detected entering the robot's workspace, and when the robot motion is stopped.
The time difference between the timestamps corresponding to these two events is recorded as the latency of our system.

\p{Instrumented Knife}
The second component we test is an instrumented knife capable of detecting contact with objects other than meat.
This is crucial in ensuring that the robot only interacts with meat at all times, avoiding undesirable interactions with foreign objects such as bone, the cutting surface, metal hooks, etc.
The instrumented knife is embedded with a pressure sensor to allow force sensing.
The robot was pre-programmed to execute a cutting trajectory over meat samples, and artificial $3$D-printed bones were deliberately placed in the robot's path to simulate unsafe contact.
We repeat this experiment $20$ times.
To evaluate the robot's performance, we measure the following quantitative metrics: first, classification accuracy defined as the percentage of times the robot is able to correctly classify the unsafe contact with the bone; and second, latency defined as the time between the initiation of contact and the robot's stopping signal.
For the latency metric, we physically record the timestamp when the knife comes into contact with the artificial bone, and when the robot comes to a complete stop.
The difference between these two timestamps serves as the latency measure in our evaluation.

\p{Uncertainty Detection}
The third component of our framework detects the uncertainty in the robot's performance during cutting tasks by analyzing the pre- and post-cutting images of the meat.
We use the movement of the meat after the cut as a proxy to assess if the cut was successfully executed.
We evaluate our uncertainty detection methodology by performing preliminary experiments that are representative of actual meat cutting operations.
We make predefined cuts on the meat.
These predefined cuts are incorrect, i.e., they are designed such that the bone intersects with the planned robot trajectory and the knife inevitably comes into contact with the meat.
Using \eq{uncertainty}, we measure the robot's uncertainty about performance.

\subsubsubsection{Full System Demonstration}
Upon verifying the performance of individual components, the final phase involves an in-person demonstration of the fully-integrated framework to evaluate its performance in real-world meat processing applications.
We collect feedback from $20$ experts, $10$ working in meat processing and $10$ in the robotics domain.
The demonstration involves processing a pork loin by first \textit{slicing} it into pork chops, then \textit{trimming} the excess fat off of the pork chops.
These tasks --- although simple --- are representative of the tasks performed in meat processing plants.
To evaluate our framework we perform the demonstration in four sections.
These sections were designed to compare our proposed framework with the industry-standard practices on two measures: \textit{safety} and \textit{performance}.
Specifically, the first two sections of the demo test the safety components of our framework --- human hand detection, the instrumented knife, and the LED interface.
The last two sections of the demo test the components that can help improve the quality of human-robot collaboration, namely the communication interface, feedback integration, and uncertainty detection. 
We name these sections as \textit{Manual Safety}, \textit{Automatic Safety}, \textit{Fully Autonomous}, and \textit{Collaborative}.
Each of these methods uses the same planning procedure for calculating the robot cutting trajectory.
The first two methods demonstrate slicing a pork loin into pork chops, and the last two method demonstrate trimming excess fat off of a pork chop.
We discuss the details of these methods below.

\p{Manual Safety} This method operates autonomously for planning the cutting trajectory to slice a pork loin into pork chops. 
Here we use manual emergency button which an operator can press to stop the robot's motion in case there is an anomaly such as the knife coming into contact with a bone or someone entering the robot's workspace.
We highlight that this is the standard practice in industry.

\p{Automatic Safety} This method uses the same planning procedure for cutting the meat.
However, we use our frameworks safety components instead of the industry-standard emergency stop button. 
Our hand detection system continuously monitors the human's movement and the instrumented knife ensures that the robot only cuts through meat. 
In case the human enters the robot's workspace or the knife makes contact with an undesirable object, the robot stops its motion and flashes the LEDs red to indicate the anomaly.

\p{Fully Autonomous} In this method, we operate the robot without maintaining transparency about its planning procedure; the robot plans and executes the cut without notifying the human about its plan.
Here we perform the trimming operation to remove excess fat from the pork chop obtained from the previous step.

\p{Collaborative} This method maintains transparency by displaying the robot's planning procedure as well as the final cutting trajectory using our graphical interface which was discussed in Section \ref{M23}. 
In addition to maintaining transparency, we also leverage the uncertainty detection component of our framework.
After the robot executes the cut, it calculates and displays its uncertainty about its performance to the human.
In case the robot is uncertain about its operation it flashes the LED red in order to inform the human that they must inspect the process and make necessary changes.

To evaluate our framework and compare it to the industry standard practices, we conduct three separate surveys. These surveys were approved by the Virginia Tech Institutional Review Board, protocol $24-736$.
First, prior to our demonstrations, we conduct a pre-demo survey to gauge the experts' perception of automation in meat processing application.
Additionally, after demonstrating each method we conduct a demo survey where we ask the experts to rate the system on three metrics: safety, less need for monitoring, and usefulness.
For each metric the experts rate the system on a Likert scale from $1$-$7$.
Finally, after demonstrating all four demonstrations, we conduct a post-demo survey to gauge how the experts' overall perception of automation in a meat processing setting has changed.
We also ask the experts to rank the methods and the different components of our framework to determine which specific parts of our framework are more important according to the industry experts.
The details of each of the three surveys are provided in Appendix \ref{appendix:survey_questions}.
\section{Results}
In this section we discuss the results of our experiments.
First, in Section \ref{sec:individual_exp} we summarize the results of our controlled experiments for evaluating the reliability of the different components.
Second, in Section \ref{sec:demos} we present the results of the demonstration we conducted to evaluate the entire framework. Videos of our experiments can be found here: \url{https://youtu.be/-DBFf64xDMs}.

\subsection{Component Evaluation}\label{sec:individual_exp}
Before using our framework for actual meat processing tasks, we conducted thorough evaluation of the individual components to ensure they can function reliably.
Specifically, we tested three main components of our framework, namely the human hand monitoring system, the instrumented knife, and the uncertainty detection module.
Each component was tested in a controlled setting where we simulate a meat processing workflow.

\begin{figure}
    \centering
    \includegraphics[width=1\linewidth]{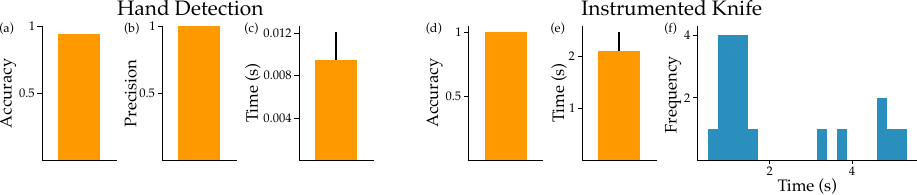}
    \caption{Results of individual component experiments. The first three plots show the results of the experiments evaluating the hand monitoring system; the last three plots show the results for the instrumented knife experiments. (a) shows the detection accuracy of our hand monitoring mechanism. (b) shows the precision of hand detection experiments defined as the ratio of the true positive predictions to the total positive predictions, and (c) shows the mean time taken by our framework to detect the human hand entering the robot's workspace and stop the robot's motion. The bar shows the standard error of mean. (d) is the accuracy of detecting the knife's contact with bone and (e) shows the mean time elapsed between the initiation of contact and stopping the robot's motion. In plot (f) we illustrate the histogram of the latency across the $20$ trials.}
    \label{fig:individual}
\end{figure}

\begin{table}[h]
\centering
\begin{tabular}{c|c|c}
\hline
                             & Movement & Uncertainty \\ \hline
\multirow{5}{*}{Translation (cm)} & 0        & 1.8 \%      \\
                             & 1.1      & 11.9 \%     \\
                             & 1.5      & 21.89 \%    \\
                             & 2.9      & 46.38 \%    \\
                             & 4.5      & 92.98 \%    \\ \hline
\multirow{5}{*}{Rotation (degrees)}    & 0        & 0.9 \%      \\
                             & 7        & 15.3 \%     \\
                             & 10       & 22.6 \%     \\
                             & 39       & 92.99 \%    \\
                             & 45       & 97.49 \%    \\ \hline
\end{tabular}
\caption{Results of the experiments evaluating the uncertainty detection module. The first five rows belong to the trials where the meat moves along the robot's trajectory and the last five rows belong to the trials where the knife rotates the meat when the knife comes into contact with the bone. The first column shows the displacement value measured manually after each execution (translation is measured in centimeters and rotation is measured in degrees). The final column shows the uncertainty value as output by our framework. The uncertainty value increases sharply with larger displacement or rotation, indicating to the operator a higher concern over the accuracy of the cut.}
\label{tab:uncertainty}
\end{table}

\p{Human Hand Monitoring System} 
The results of our human hand monitoring system tests are illustrated in the first three plots in \fig{individual}.
Our system achieved an accuracy of $96 \%$, correctly detecting the presence of human hands inside the robot's workspace $47$ times out of $50$ trials, as shown in the first plot from the left.
Notably, the system achieved no false positives, i.e., our framework did not falsely detect the human hand when the human is outside the workspace, thereby achieving a precision of $1$.
However, there were three instances of false negatives where the system failed to detect the human hand.
The mean latency across all trials was $0.0095$s, with a standard deviation of $0.0026$s.

\p{Instrumented Knife} Next, we tested the reliability of the instrumented knife to detect contact with objects other than meat and stopping the robot motion accordingly.
The three plots on the right in \fig{individual} summarize our results across $20$ trials.
The first plot shows the accuracy, the second plot shows the mean time our framework took to detect the unsafe contact and stop the robot's motion, and the third plot shows a histogram of the times across $20$ trials.
We see that our framework achieved $100 \%$ accuracy, correctly detecting the contact every time.
Further, the mean time lag of our framework was $2.095$s, with a standard deviation of $1.62$s.
We note that, $75 \%$ of the trials took less than $2$ seconds.
The higher mean is likely a result of the three trials that took more than $4$ seconds.

\p{Uncertainty Detection} Finally, we tested the uncertainty detection mechanism of our framework in a controlled setting.
We performed predefined cutting operations where the bone intersected with the robot's trajectory.
As a result, the robot was unable to successfully complete the cut and instead dragged the meat along its trajectory.
Here we report the uncertainty calculated by our framework.
The results are summarized in Table \ref{tab:uncertainty}. 
We see that the uncertainty value steeply increased the more the meat moved.
Slight movements of meat resulted in lower uncertainty; this makes sense, because this movement could simply be the result of the meat deforming after a successful cut.
However, the uncertainty value was significantly larger in proportion to the magnitude of the meat's movement, highlighting situations where the cut was most likely erroneous.
Interestingly, our framework was able to detect translational as well as rotational movement of meat.

\subsection{Overall Framework Evaluation}\label{sec:demos}

\begin{figure}
    \centering
    \includegraphics[width=0.75\linewidth]{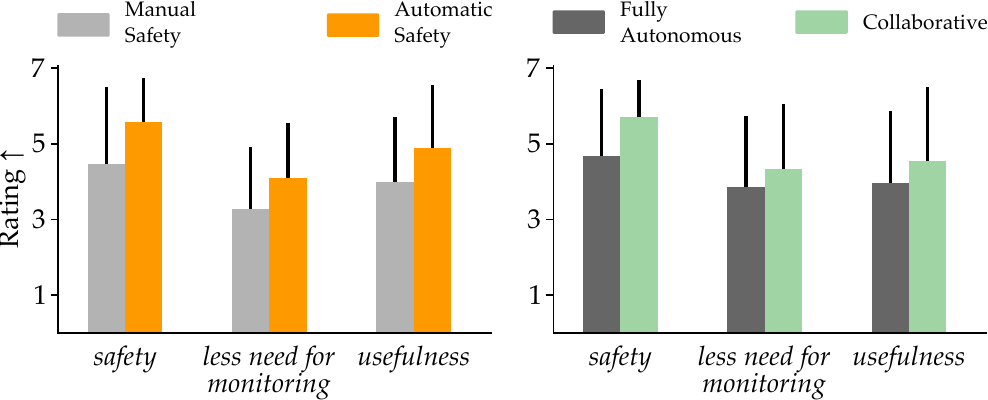}
    \caption{Results of the survey conducted after each of the four sections. The experts rated these on three measures, namely \textit{safety}, \textit{less need for monitoring}, and \textit{usefulness}. A higher value indicates a more favorable rating. (Left) The plot shows the participants' ratings for the first two methods: Manual Safety where the robot setup uses the industry standard emergency stop button to stop the robot and Automatic Safety which uses our framework's safety mechanism including the hand monitoring system and the instrumented knife. (Right) The results of the demo survey for the final two methods: Fully Autonomous where the robot plans and executes the cut as a black-box without communicating with the human and Collaborative where the robot uses our framework's communication and feedback integration mechanisms during planning.}
    \label{fig:likert}
\end{figure}

To evaluate our entire framework, we performed a demonstration where the robot performed two meat processing applications --- slicing a pork loin into pork chops, and trimming the excess fat off the pork chops.
Specifically, we divided the demos into four sections where we compared the industry standard approaches with our proposed approach.
These four sections are Manual Safety, Automatic Safety, Fully Autonomous, and Collaborative as detailed in Section \ref{sec:evaluation_procedure}.
In \fig{likert} we illustrate the results of the demo survey administered to the experts after each demonstrations asking them to rate our framework on three metrics: \textit{safety}, \textit{less need for monitoring}, and \textit{usefulness}.
A higher value for these metrics indicates the experts rated it more positively.
We can see that across all three metrics the experts rated the sections of the demo that use our framework, namely Automatic Safety and Collaborative, more favorably than the sections that use the industry standard for automation.
We performed the non-parametric Friedman test to measure statistical significance. We found that there was a statistically significant difference in the ratings for \textit{safety} ($\chi^2=11.012$, $p=0.012$), \textit{less need for monitoring} ($\chi^2=10.174$, $p=0.017$), and \textit{usefulness} ($\chi^2=12.210$, $p=0.007$).

\begin{figure}
    \centering
    \includegraphics[width=1\linewidth]{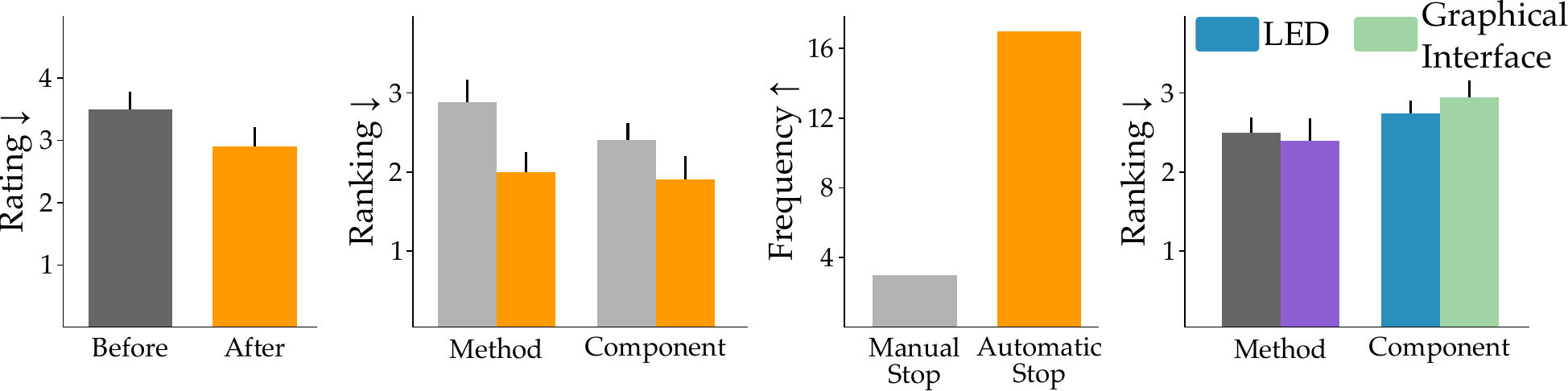}
    \caption{Results of the post-demo survey. (Left) The plot shows how the experts' perception about safety changes after witnessing our framework compared to pre-demonstration results. (Middle-left) The first bar plot shows how the experts ranked the Manual Safety and Automatic safety methods as a whole. The second bar plot shows how the expert ranking of the individual components, i.e., emergency stop button and our framework's safety mechanisms. (Middle-right) The experts were asked to choose between the industry standard manual safety stop button and our framework which automatically stops the robot in case of danger. (Right) Participants compared the Fully Autonomous method and the Collaborative method by ranking them relative to each other. Additionally, they also ranked the individual components of our framework --- LED interface and the Transparency interfaces --- to indicate which component was more favorable in practice.}
    \label{fig:pre_post}
\end{figure}

We also measured how the experts' perception of cobots in meat processing applications changed after attending the demonstrations through pre- and post-demo surveys. 
The results are presented in \fig{pre_post}.
The plot on the left indicates how the experts' concern for safety changed after observing the demonstration of our framework.
A lower value indicates that they were less concerned about safety.
It is evident that the experts, on average, reported a lower concern after observing the demonstration.
Although a one-sided paired t-test revealed that the results were not statistically significant ($p=0.052$), it did indicate a marginal trend towards reduced concern for safety following the demonstration.
The second plot shows how the experts ranked the first two sections of the demo which compare the safety components of our framework.
Additionally, the experts were also asked to explicitly rate the safety components --- manual emergency stop button and our hand detection module.
A lower value indicates that the experts ranked it more favorably (i.e., they would prioritize having this capability).
From the plot it is evident that the experts preferred our framework compared to the standard approach.
The third plot shows that out of the $20$ experts, $17$ preferred the automatic stop mechanism.
Finally, the experts were also asked to rank the final two sections of the demo, namely Fully Autonomous and Collaborative approaches.
They were also asked to rank the transparency components based on their usefulness.
We see that the mean ranking of the two sections of the demo are similar.
Along the same lines, participants seemed to have an equal preference for LED and graphical interface.

\section{Discussion}

In our experiments we assessed the main components of our framework in a controlled environment.
First, we tested the accuracy and responsiveness of the human hand-monitoring system.
Specifically, we tested how reliably our framework could identify instances when the human hand enters the robot's workspace and how quickly it could stop the robot's motion in response.
The results indicate that our system achieved a high accuracy with a latency in the order of milliseconds.
Such quick response is crucial for ensuring operational safety, particularly in environments like meat processing facilities where close human-robot interaction is inevitable.
While our current approach focused on monitoring the human hand, a more comprehensive approach would involve tracking the entire human body \citep{qiao2017real,guler2018densepose,redmon2016you}.
However, we argue that this presents significant computational challenges.
Commonly available human detection models are resource-intensive and often require high-power GPUs for inference, making them impractical for deployment in meat processing environments that lack the infrastructure for such hardware.
In contrast, our framework employed a lightweight Mediapipe model, which was able to achieve real-time inference on small hardware such as the intel NUC.
Additionally, the Mediapipe model exhibits robustness under various lighting conditions, though in meat processing facilities consistent illumination can be assumed.
Despite these advantages, there remain several avenues for future research.
One possible direction is the integration of active collision avoidance strategies.
While the current framework stops the robot's movement to minimize the risk of collision, future iterations could integrate adaptive planning based on efficient scene representation, allowing the robot to re-plan its trajectories dynamically in the presence of humans \citep{ali2024interactive}.
Moreover, the presence of human can also be used to turn off the knife through an actuated knife mount.

Next, we evaluated the performance of the instrumented knife in detecting contact with foreign objects that differ in their physical properties from meat.
Our findings indicate that the system reliably detected contact with harder surfaces, such as bone, and stopped the robot's motion to prevent potential hazards.
The observed high latency in the framework's response can be attributed to the knife sliding over the bone; this can be easily mitigated by tuning the force threshold.
A more robust alternative involves developing specialized knives equipped with integrated sensors designed to detect and classify contact with various objects \citep{aly2024tactile,aly2025fundamental}.
While this approach is technologically superior, its cost effectiveness must be carefully considered given the constraints of small and medium scale meat processing industries.

Further, we tested our framework's capability at detecting uncertainty, which is crucial in collaborative human-robot scenarios.
Robots must be able to assess their performance and communicate anomalies to human collaborators when necessary.
In our framework we used displacement of meat as a proxy measure for error in the robot's cuts.
Experimental results suggested that when the knife fails to cut through the bone, it may displace the meat while the robot continues along its planned trajectory.
Thus, displacement can serve as a viable indicator for anomalies.
We acknowledge that more sophisticated machine learning approaches could be applied to enhance uncertainty detection.
Large vision language models (VLMs) can potentially be leveraged for identifying errors in meat processing tasks.
However, we believe that such models may not be applicable out-of-the-box and may need fine-tuning on meat processing-specific datasets.
Future research efforts could be directed towards developing meat processing datasets or developing vision based models capable of comparing images of the meat before and after the cut for estimating the accuracy. 

Following these component level experiments, we conducted a demonstration involving experts from robotics and meat-processing to qualitatively evaluate the safety and usefulness of our framework.
The demonstration was divided into four sections: the first two compared the safety mechanisms with the existing industry standard practices, while the latter two compared the performance-oriented features, i.e., transparency and communication, against the conventional fully autonomous operation.
The survey results clearly indicated a preference for our proposed framework over the current industrial standard.
The experts noted that our framework was safer, required less human oversight, and consisted of features that were particularly suited for meat processing operation.
It is noted that the post-demo survey suggested a positive shift in the overall perception of the experts regarding the safety of automation in meat processing.
This downward trend in their concern points to the potential for such a framework to enhance the viability of cobots in meat industry.
Overall, the experts emphasized safety as the top-most priority when considering robots in meat processing plants.
This is entirely justified given the inherent risks of working in close proximity of a robot equipped with cutting tools.
Therefore, future research should focus on improving the safety of collaborative robots.
Although additional elements of the framework such as graphical and LED interfaces were deemed useful, they were considered secondary to the critical safety functionality.
The almost unanimous preference for automatic stop over manual stop underscores the need for developing proactive safety measures.

\section{Conclusion}

In this work we developed and tested a safety and transparency framework to facilitate integrating collaborative robots into the meat processing industry.
Our framework addresses three main challenges we identified through a survey of meat industry experts: \textit{safety}, \textit{transparency}, and \textit{feedback integration}.
To ensure safety, we added two components to our general-purpose robot arm.
First, a human hand detection system that continuously monitors the human relative to the robot in order to reduce the risk of collision with humans.
Second, an instrumented knife that can determine when the robot is cutting through the intended meat, and when the robot collides with an unintended object (e.g., a bone or fixture in the environment).
In order to maintain transparency during the robot's operation, we leverage two interfaces for communicating the robot's state and planning process.
An LED interface uses lights to indicate any anomalies in the robot's operation such as the human coming too close to the robot, and a graphical interface that displays the robot's planning process to the human coworkers.
This graphical interface also allows the human to provide feedback on the robot's planned cut through a click-and-drag operation; workers can edit the robot's displayed plan as needed.

We evaluate each of the components of our framework separately through controlled experiments.
The results suggest that these components can reliably ensure safety and transparency.
Additionally, we perform a demonstration where we process pork-loins by slicing them into chops and removing the fat from those chops.
These demonstrations compare our framework to the standard tools that are used in industrial automation.
For evaluation, we invited $20$ experts from robotics and meat processing to provide their subjective assessment of the two frameworks.
The expert feedback suggests that our novel framework represents a step forward in integrating collaborative robots into the meat processing workflow.
By enhancing human and operational safety as well as introducing transparency into the robot's workflow, these frameworks have the potential to improve job satisfaction, productivity, and supply chain consistency.

\bibliography{citations}

\begin{thebibliography}{10}
\urlstyle{rm}
\expandafter\ifx\csname url\endcsname\relax
  \def\url#1{\texttt{#1}}\fi
\expandafter\ifx\csname urlprefix\endcsname\relax\def\urlprefix{URL }\fi
\expandafter\ifx\csname doiprefix\endcsname\relax\def\doiprefix{DOI: }\fi
\providecommand{\bibinfo}[2]{#2}
\providecommand{\eprint}[2][]{\url{#2}}

\bibitem{fao}
\bibinfo{author}{Food} \& \bibinfo{author}{Organization, A.}
\newblock \bibinfo{title}{Fao} (\bibinfo{year}{2024}).
\newblock \bibinfo{note}{\url{https://openknowledge.fao.org/handle/20.500.14283/cd5077en}}.

\bibitem{taylor2020}
\bibinfo{author}{Taylor, C.~A.}, \bibinfo{author}{Boulos, C.} \& \bibinfo{author}{Almond, D.}
\newblock \bibinfo{journal}{\bibinfo{title}{Livestock plants and {COVID}-19 transmission}}.
\newblock {\emph{\JournalTitle{Proceedings of the National Academy of Sciences}}}  (\bibinfo{year}{2020}).

\bibitem{bir2021impact}
\bibinfo{author}{Bir, C.}, \bibinfo{author}{Peel, D.}, \bibinfo{author}{Holcomb, R.}, \bibinfo{author}{Raper, K.} \& \bibinfo{author}{Jones, J.}
\newblock \bibinfo{title}{The impact of covid-19 on meat processing, and the renewed interest in local processing capabilities}.
\newblock In \emph{\bibinfo{booktitle}{Western Economics Forum}} (\bibinfo{year}{2021}).

\bibitem{victor2016slaughtering}
\bibinfo{author}{Victor, K.} \& \bibinfo{author}{Barnard, A.}
\newblock \bibinfo{journal}{\bibinfo{title}{Slaughtering for a living: A hermeneutic phenomenological perspective on the well-being of slaughterhouse employees}}.
\newblock {\emph{\JournalTitle{International journal of qualitative studies on health and well-being}}}  (\bibinfo{year}{2016}).

\bibitem{gaston2012meatpacking}
\bibinfo{author}{Gast{\'o}n, M.} \& \bibinfo{author}{Harrison, W.}
\newblock \bibinfo{journal}{\bibinfo{title}{Meatpacking workers' perceptions of working conditions, psychological contracts, and organizational justice}}.
\newblock {\emph{\JournalTitle{Journal of Latino/Latin American Studies}}}  (\bibinfo{year}{2012}).

\bibitem{pastrana2023slaughterhouse}
\bibinfo{author}{Pastrana-Camacho, A.~P.}, \bibinfo{author}{Est{\'e}vez-Moreno, L.~X.} \& \bibinfo{author}{Miranda-de~la Lama, G.~C.}
\newblock \bibinfo{journal}{\bibinfo{title}{What slaughterhouse workers' attitudes and knowledge reveal about human-pig relationships during pre-slaughter operations: A profile-based approach}}.
\newblock {\emph{\JournalTitle{Meat science}}}  (\bibinfo{year}{2023}).

\bibitem{slade2023psychological}
\bibinfo{author}{Slade, J.} \& \bibinfo{author}{Alleyne, E.}
\newblock \bibinfo{journal}{\bibinfo{title}{The psychological impact of slaughterhouse employment: A systematic literature review}}.
\newblock {\emph{\JournalTitle{Trauma, Violence, \& Abuse}}}  (\bibinfo{year}{2023}).

\bibitem{aly2023robotics}
\bibinfo{author}{Aly, B.~A.}, \bibinfo{author}{Low, T.}, \bibinfo{author}{Long, D.}, \bibinfo{author}{Baillie, C.} \& \bibinfo{author}{Brett, P.}
\newblock \bibinfo{journal}{\bibinfo{title}{Robotics and sensing technologies in red meat processing: A review}}.
\newblock {\emph{\JournalTitle{Trends in Food Science \& Technology}}}  (\bibinfo{year}{2023}).

\bibitem{madsen2006automation}
\bibinfo{author}{Madsen, N.}, \bibinfo{author}{Nielsen, J.~U.} \& \bibinfo{author}{M{\o}nsted, J.}
\newblock \bibinfo{title}{Automation--the meat factory of the future}.
\newblock In \emph{\bibinfo{booktitle}{52nd International Congress of Meat Science and Technology}} (\bibinfo{year}{2006}).

\bibitem{wakholi2021economic}
\bibinfo{author}{Wakholi, C.} \emph{et~al.}
\newblock \bibinfo{journal}{\bibinfo{title}{Economic analysis of an image-based beef carcass yield estimation system in korea}}.
\newblock {\emph{\JournalTitle{Animals}}}  (\bibinfo{year}{2021}).

\bibitem{kim2021economic}
\bibinfo{author}{Kim, J.} \emph{et~al.}
\newblock \bibinfo{journal}{\bibinfo{title}{Economic analysis of the use of vcs2000 for pork carcass meat yield grading in korea}}.
\newblock {\emph{\JournalTitle{Animals}}}  (\bibinfo{year}{2021}).

\bibitem{hinrichsen2010manufacturing}
\bibinfo{author}{Hinrichsen, L.}
\newblock \bibinfo{journal}{\bibinfo{title}{Manufacturing technology in the danish pig slaughter industry}}.
\newblock {\emph{\JournalTitle{Meat science}}}  (\bibinfo{year}{2010}).

\bibitem{kakade2023applications}
\bibinfo{author}{Kakade, S.}, \bibinfo{author}{Patle, B.} \& \bibinfo{author}{Umbarkar, A.}
\newblock \bibinfo{journal}{\bibinfo{title}{Applications of collaborative robots in agile manufacturing: a review}}.
\newblock {\emph{\JournalTitle{Robotic Systems and Applications}}}  (\bibinfo{year}{2023}).

\bibitem{keshvarparast2024collaborative}
\bibinfo{author}{Keshvarparast, A.}, \bibinfo{author}{Battini, D.}, \bibinfo{author}{Battaia, O.} \& \bibinfo{author}{Pirayesh, A.}
\newblock \bibinfo{journal}{\bibinfo{title}{Collaborative robots in manufacturing and assembly systems: literature review and future research agenda}}.
\newblock {\emph{\JournalTitle{Journal of Intelligent Manufacturing}}}  (\bibinfo{year}{2024}).

\bibitem{parekh2023learning}
\bibinfo{author}{Parekh, S.} \& \bibinfo{author}{Losey, D.~P.}
\newblock \bibinfo{journal}{\bibinfo{title}{Learning latent representations to co-adapt to humans}}.
\newblock {\emph{\JournalTitle{Autonomous Robots}}}  (\bibinfo{year}{2023}).

\bibitem{puttero2025collaborative}
\bibinfo{author}{Puttero, S.}, \bibinfo{author}{Verna, E.}, \bibinfo{author}{Genta, G.} \& \bibinfo{author}{Galetto, M.}
\newblock \bibinfo{journal}{\bibinfo{title}{Collaborative robots for quality control: an overview of recent studies and emerging trends}}.
\newblock {\emph{\JournalTitle{Journal of Intelligent Manufacturing}}}  (\bibinfo{year}{2025}).

\bibitem{romanov2022towards}
\bibinfo{author}{Romanov, D.}, \bibinfo{author}{Korostynska, O.}, \bibinfo{author}{Lekang, O.~I.} \& \bibinfo{author}{Mason, A.}
\newblock \bibinfo{journal}{\bibinfo{title}{Towards human-robot collaboration in meat processing: Challenges and possibilities}}.
\newblock {\emph{\JournalTitle{Journal of Food Engineering}}}  (\bibinfo{year}{2022}).

\bibitem{sahan2023role}
\bibinfo{author}{Sahan, A.~M.}, \bibinfo{author}{Kathiravan, S.}, \bibinfo{author}{Lokesh, M.} \& \bibinfo{author}{Raffik, R.}
\newblock \bibinfo{title}{Role of cobots over industrial robots in industry 5.0: A review}.
\newblock In \emph{\bibinfo{booktitle}{2023 2nd International Conference on Advancements in Electrical, Electronics, Communication, Computing and Automation (ICAECA)}} (\bibinfo{year}{2023}).

\bibitem{fournier2024human}
\bibinfo{author}{Fournier, {\'E}.} \emph{et~al.}
\newblock \bibinfo{journal}{\bibinfo{title}{Human-cobot collaboration's impact on success, time completion, errors, workload, gestures and acceptability during an assembly task}}.
\newblock {\emph{\JournalTitle{Applied Ergonomics}}}  (\bibinfo{year}{2024}).

\bibitem{bouillet2025effects}
\bibinfo{author}{Bouillet, K.}, \bibinfo{author}{Lemonnier, S.}, \bibinfo{author}{Clanche, F.} \& \bibinfo{author}{Gauchard, G.}
\newblock \bibinfo{journal}{\bibinfo{title}{Effects of pace on productivity and physical and mental workloads in a human--cobot collaboration}}.
\newblock {\emph{\JournalTitle{International Journal of Occupational Safety and Ergonomics}}}  (\bibinfo{year}{2025}).

\bibitem{su2024exploring}
\bibinfo{author}{Su, B.} \emph{et~al.}
\newblock \bibinfo{journal}{\bibinfo{title}{Exploring the impact of human-robot interaction on workers' mental stress in collaborative assembly tasks}}.
\newblock {\emph{\JournalTitle{Applied Ergonomics}}}  (\bibinfo{year}{2024}).

\bibitem{wright2024safely}
\bibinfo{author}{Wright, R.}, \bibinfo{author}{Parekh, S.}, \bibinfo{author}{White, R.} \& \bibinfo{author}{Losey, D.~P.}
\newblock \bibinfo{journal}{\bibinfo{title}{Safely and autonomously cutting meat with a collaborative robot arm}}.
\newblock {\emph{\JournalTitle{Scientific Reports}}}  (\bibinfo{year}{2024}).

\bibitem{xing2018smart}
\bibinfo{author}{Xing, B.} \& \bibinfo{author}{Marwala, T.}
\newblock \bibinfo{journal}{\bibinfo{title}{Smart maintenance for human--robot interaction}}.
\newblock {\emph{\JournalTitle{Studies in Systems, Decision and Control. Springer}}}  (\bibinfo{year}{2018}).

\bibitem{mutlu2016cognitive}
\bibinfo{author}{Mutlu, B.}, \bibinfo{author}{Roy, N.} \& \bibinfo{author}{{\v{S}}abanovi{\'c}, S.}
\newblock \bibinfo{journal}{\bibinfo{title}{Cognitive human--robot interaction}}.
\newblock {\emph{\JournalTitle{Springer handbook of robotics}}}  (\bibinfo{year}{2016}).

\bibitem{wright2025survey}
\bibinfo{author}{Wright, R.}, \bibinfo{author}{Parekh, S.}, \bibinfo{author}{Losey, D.~P.} \& \bibinfo{author}{White, R.}
\newblock \bibinfo{title}{Surveying processor perceptions of automation in meat processing} (\bibinfo{year}{2025}).
\newblock \bibinfo{note}{Under review}.

\bibitem{haddadin2016physical}
\bibinfo{author}{Haddadin, S.} \& \bibinfo{author}{Croft, E.}
\newblock \bibinfo{title}{Physical human--robot interaction}.
\newblock In \emph{\bibinfo{booktitle}{Springer handbook of robotics}}, \bibinfo{pages}{1835--1874} (\bibinfo{publisher}{Springer}, \bibinfo{year}{2016}).

\bibitem{lasota2017survey}
\bibinfo{author}{Lasota, P.~A.}, \bibinfo{author}{Fong, T.} \& \bibinfo{author}{Shah, J.~A.}
\newblock \bibinfo{journal}{\bibinfo{title}{A survey of methods for safe human-robot interaction}}.
\newblock {\emph{\JournalTitle{Foundations and Trends in Robotics}}}  (\bibinfo{year}{2017}).

\bibitem{habibian2025survey}
\bibinfo{author}{Habibian, S.}, \bibinfo{author}{Alvarez~Valdivia, A.}, \bibinfo{author}{Blumenschein, L.~H.} \& \bibinfo{author}{Losey, D.~P.}
\newblock \bibinfo{journal}{\bibinfo{title}{A survey of communicating robot learning during human-robot interaction}}.
\newblock {\emph{\JournalTitle{The International Journal of Robotics Research}}}  (\bibinfo{year}{2025}).

\bibitem{yang2017evaluating}
\bibinfo{author}{Yang, X.~J.}, \bibinfo{author}{Unhelkar, V.~V.}, \bibinfo{author}{Li, K.} \& \bibinfo{author}{Shah, J.~A.}
\newblock \bibinfo{title}{Evaluating effects of user experience and system transparency on trust in automation}.
\newblock In \emph{\bibinfo{booktitle}{Proceedings of the 2017 ACM/IEEE international conference on human-robot interaction}} (\bibinfo{year}{2017}).

\bibitem{yao2025robot}
\bibinfo{author}{Yao, M.}, \bibinfo{author}{Li, J.} \& \bibinfo{author}{Wang, Z.}
\newblock \bibinfo{journal}{\bibinfo{title}{Robot transparency and employees’ acceptance: The roles of trust and anthropomorphism}}.
\newblock {\emph{\JournalTitle{International Journal of Human--Computer Interaction}}}  (\bibinfo{year}{2025}).

\bibitem{de2008atlas}
\bibinfo{author}{De~Santis, A.}, \bibinfo{author}{Siciliano, B.}, \bibinfo{author}{De~Luca, A.} \& \bibinfo{author}{Bicchi, A.}
\newblock \bibinfo{journal}{\bibinfo{title}{An atlas of physical human--robot interaction}}.
\newblock {\emph{\JournalTitle{Mechanism and Machine Theory}}}  (\bibinfo{year}{2008}).

\bibitem{mediapipe}
\bibinfo{author}{Lugaresi, C.} \emph{et~al.}
\newblock \bibinfo{title}{Mediapipe: A framework for perceiving and processing reality}.
\newblock In \emph{\bibinfo{booktitle}{Third Workshop on Computer Vision for AR/VR at IEEE Computer Vision and Pattern Recognition (CVPR) 2019}} (\bibinfo{year}{2019}).

\bibitem{flexiforce}
\bibinfo{author}{Tekscan}.
\newblock \bibinfo{title}{Flexiforce a401 sensor}.
\newblock \bibinfo{note}{\url{https://www.tekscan.com/products-solutions/force-sensors/flexiforce-a401-sensor}}.

\bibitem{winfield2021ieee}
\bibinfo{author}{Winfield, A.~F.} \emph{et~al.}
\newblock \bibinfo{journal}{\bibinfo{title}{Ieee p7001: A proposed standard on transparency}}.
\newblock {\emph{\JournalTitle{Frontiers in Robotics and AI}}}  (\bibinfo{year}{2021}).

\bibitem{ur10}
\bibinfo{author}{UniversalRobots}.
\newblock \bibinfo{title}{{UR10 A}pplications}.
\newblock \bibinfo{note}{\url{https://www.universal-robots.com/applications/}}.

\bibitem{ur10_2}
\bibinfo{author}{UniversalRobots}.
\newblock \bibinfo{title}{{UR10 C}ompliance}.
\newblock \bibinfo{note}{\url{https://www.universal-robots.com/articles/ur/safety/safety-faq/}}.

\bibitem{qiao2017real}
\bibinfo{author}{Qiao, S.}, \bibinfo{author}{Wang, Y.} \& \bibinfo{author}{Li, J.}
\newblock \bibinfo{title}{Real-time human gesture grading based on openpose}.
\newblock In \emph{\bibinfo{booktitle}{2017 10th International Congress on Image and Signal Processing, BioMedical Engineering and Informatics (CISP-BMEI)}} (\bibinfo{year}{2017}).

\bibitem{guler2018densepose}
\bibinfo{author}{G{\"u}ler, R.~A.}, \bibinfo{author}{Neverova, N.} \& \bibinfo{author}{Kokkinos, I.}
\newblock \bibinfo{title}{Densepose: Dense human pose estimation in the wild}.
\newblock In \emph{\bibinfo{booktitle}{Proceedings of the IEEE conference on computer vision and pattern recognition}} (\bibinfo{year}{2018}).

\bibitem{redmon2016you}
\bibinfo{author}{Redmon, J.}, \bibinfo{author}{Divvala, S.}, \bibinfo{author}{Girshick, R.} \& \bibinfo{author}{Farhadi, A.}
\newblock \bibinfo{title}{You only look once: Unified, real-time object detection}.
\newblock In \emph{\bibinfo{booktitle}{Proceedings of the IEEE conference on computer vision and pattern recognition}} (\bibinfo{year}{2016}).

\bibitem{ali2024interactive}
\bibinfo{author}{Ali, U.} \emph{et~al.}
\newblock \bibinfo{journal}{\bibinfo{title}{Interactive distance field mapping and planning to enable human-robot collaboration}}.
\newblock {\emph{\JournalTitle{IEEE Robotics and Automation Letters}}}  (\bibinfo{year}{2024}).

\bibitem{aly2024tactile}
\bibinfo{author}{Aly, B.~A.}, \bibinfo{author}{Low, T.}, \bibinfo{author}{Long, D.}, \bibinfo{author}{Brett, P.} \& \bibinfo{author}{Baillie, C.}
\newblock \bibinfo{journal}{\bibinfo{title}{Tactile sensing for tissue discrimination in robotic meat cutting: A feasibility study}}.
\newblock {\emph{\JournalTitle{Journal of Food Engineering}}}  (\bibinfo{year}{2024}).

\bibitem{aly2025fundamental}
\bibinfo{author}{Aly, B.~A.}, \bibinfo{author}{Brett, P.}, \bibinfo{author}{Low, T.} \& \bibinfo{author}{Long, D.}
\newblock \bibinfo{journal}{\bibinfo{title}{Fundamental studies on tactile feedback in robotic striploin fat trimming task}}.
\newblock {\emph{\JournalTitle{Journal of Food Engineering}}}  (\bibinfo{year}{2025}).

\end{thebibliography}

\section{Appendix}

\subsection{Survey Questions}\label{appendix:survey_questions}
In this section we provide the details of the questions we asked the participants in our demonstration.
We developed a total of three surveys.
Prior to the demonstrations, the participants completed a pre-demo survey (Table \ref{tab:pre-demo}) designed to collect background information and assess their initial perception about automation in meat processing.
As outlined in Section \ref{sec:evaluation_procedure}, the demonstrations consisted of four methods: Manual Safety, Automatic Safety, Fully Autonomous, and Collaborative.
Following each method, the participants completed a demo survey, shown in Table \ref{tab:demo}, which evaluated the performance quality of the respective method.
Finally, after all the demonstrations participants completed a post-demo survey outlined in Table \ref{tab:post-demo} aimed at capturing their overall perception of automation in meat processing.
This survey also included questions regarding the perceived usefulness of individual components of our framework.
The tables below list the questions pertaining to each of these three surveys. 

\begin{table}[h]
    \centering
    \begin{tabular}{p{13cm}|m{3cm}}
    \hline
        Question & Response Type \\
    \hline
        How familiar are you with automated meat processing & scale of 1-5 \\
        How often do you interact with automated meat processing systems? & scale from 1-4 \\
        As an individual in the meat processing industry, how concerned are you about safety in a collaborative meat processing setting? & scale of 1-5 \\
        Please detail any specific justification for your above response. & open-ended \\
    \hline
    \end{tabular}
    \caption{The questions from our pre-demo survey. Before the demonstration we administered a survey to gather data on the experts' perception of safety of automation in meat processing. The survey also included questions about their familiarity with automated meat processing to assess how well they could evaluate a robot's actions in the absence of transparency.}
    \label{tab:pre-demo}
\end{table}


\begin{table}[h]
    \centering
    \begin{tabular}{p{13cm}|m{3cm}}
    \hline
        Question & Response Type \\
    \hline
        I felt safe around the robot. & scale of 1-7 \\
        I felt nervous around the robot. & scale of 1-7 \\
        I felt that I had to monitor the robot. & scale of 1-7 \\
        I did not feel that I had to monitor the robot. & scale of 1-7 \\
        Please rank, on a scale of 1 - 7, the perceived usefulness of this system in a meat processing setting. & scale of 1-7 \\
        \hline
    \end{tabular}
    \caption{The questions asked in our demo survey. After demonstrating each of the four methods, we conducted a survey asking the experts to rate the robot's performance on three measures: safety, less need for monitoring, and usefulness. The experts rated the robot on a Likert scale from $1$ to $7$.}
    \label{tab:demo}
\end{table}


\begin{table}[h]
    \centering
    \begin{tabular}{p{13cm}|m{3cm}}
    \hline
        Question & Response Type \\
    \hline
        As an individual in the meat processing industry, how concerned are you about safety in a collaborative meat processing setting after interacting with the robot? & scale of 1-5 \\
        Please detail any specific justification for your above response. & open-ended \\
        Which safety feature modality felt safest during your interactions? & Multiple Choice \\
        Please detail any specific justification for your above response. & open-ended \\
        Please rank the following in order of their capacity to ensure task proficiency & ranking \\
        Please rank the following features in order of importance in a meat processing setting & ranking \\
        Please detail any specific justification for your above response. & open-ended \\
    \hline
    \end{tabular}
    \caption{Questions asked in the post-demo survey. At the end of the demonstrations we administered a survey to assess how the experts' perception about the robot's safety in meat processing changed. Additionally, the experts were asked to rank the methods as well as each of the individual components based on their usefulness in meat processing application. This allowed us to subjectively evaluate each of the components of our framework.}
    \label{tab:post-demo}
\end{table}

\vspace{2cm}
\section*{Conflict of Interest}
None. The authors declare no competing interests.

\section*{Funding}
This work was supported in part by the USDA National Institute of Food and Agriculture [Grant Number 2022-67021-37868].

\section*{Author contributions statement}
S.P.: Conceptualization, Investigation, Data Curation, Methodology, Formal analysis, Writing - original draft.
C.G.: Conceptualization, Investigation, Software, Validation, Writing - original draft.
R.W.: Conceptualization, Investigation, Validation, Methodology, Writing - review and editing.
R.W.: Conceptualization, Supervision, Funding Acquisition, Writing - review and editing.
D.L.: Conceptualization, Supervision, Funding Acquisition, Writing - review and editing.

\section*{Data Availability}
Data will be made available on request. Please contact Sagar Parekh (sagarp@vt.edu) to request this data.

\end{document}